\documentclass[fleqn,10pt]{wlscirep}
\usepackage[utf8]{inputenc}
\usepackage[T1]{fontenc}
\title{Architecting Large Action Models for Human-in-the-Loop Intelligent Robots}

\author{Kanisorn Sangchai, Methasit Boonpun, Withawin Kraipetchara, Paulo Garcia}
\affil{International School of Engineering, Chulalongkorn University, Bangkok, Thailand}
\affil{{6538020621,6538165021,6538191221,paulo.g}@chula.ac.th}

\begin{abstract}
The realization of intelligent robots, operating autonomously and interacting with other intelligent agents, human or artificial, requires the integration of environment perception, reasoning, and action.
Classic Artificial Intelligence techniques for this purpose, focusing on \textit{symbolic} approaches, have long-ago hit the scalability wall on compute and memory costs. Advances in Large Language Models in the past decade (\textit{neural} approaches) have resulted in unprecedented displays of capability, at the cost of control, explainability, and interpretability.
Large Action Models aim at extending Large Language Models to encompass the full perception, reasoning, and action cycle; however, they typically require substantially more comprehensive training and suffer from the same deficiencies in reliability.
Here, we show it is possible to build competent Large Action Models by composing off-the-shelf foundation models, and that their control, interpretability, and explainability can be effected by incorporating symbolic wrappers and associated verification on their outputs, achieving verifiable \textit{neuro-symbolic} solutions for intelligent robots.
Our experiments on a multi-modal robot demonstrate that Large Action Model intelligence does not require massive end-to-end training, but can be achieved by integrating efficient perception models with a logic-driven core. We find that driving action execution through the generation of Planning Domain Definition Language (PDDL) code enables a human-in-the-loop verification stage that effectively mitigates action hallucinations.
These results can support practitioners in the design and development of robotic Large Action Models across novel industries, and shed light on the ongoing challenges that must be addressed to ensure safety in the field.
\end{abstract}
\begin{document}

\flushbottom
\maketitle
\thispagestyle{empty}

\section{Introduction}

Intelligent robots capable of perceiving their surroundings, reasoning about tasks, and executing reliable actions are increasingly essential in engineering, manufacturing, and everyday collaborative settings \cite{varlamov2021brains}. Their potential impact lies not only in automating repetitive labor, but in enabling fluid human-robot collaboration where robots understand user intent, adapt to changing environments, and provide transparent, verifiable reasoning. Achieving this, however, requires solving the longstanding integration problem: connecting perception, language, symbolic reasoning, and physical control into a coherent system \cite{qiao2021survey,gan2024embodied}.

\par Classic \textit{symbolic} AI approaches offer strong guarantees on correctness, interpretability, and verification, but scale poorly with the complexity and variability of real-world environments~\cite{tamp, recent-trends-in-tamp}. Conversely, \textit{neural} methods, particularly Large Language Models (LLMs) and Vision-Language Models (VLMs), excel at perception, abstraction, and generalization, but struggle with controllability, reliability, and the prevention of hallucinations~\cite{eval-application-challenges-llms}. These limitations hinder their deployment in safety-critical robotic systems, where incorrect reasoning can lead to dangerous actions~\cite{plangenllm}.

\par Large Action Models (LAMs) have recently emerged as a potential solution to this integration gap. Defined as agents trained on large-scale offline datasets via causal sequence modeling, LAMs aim to extend the generalization capabilities of LLMs from text generation to active physical task completion~\cite{lram}. However, relying on massive end-to-end models presents distinct challenges. First, existing LAMs are primarily based on Transformer architectures, which suffer from quadratic inference complexity; this high latency is often prohibitive for real-time robotic control~\cite{lram}. Second, unlike textual hallucinations, errors in action execution ("action hallucinations") can lead to irreversible physical damage and safety hazards~\cite{lam}. Consequently, a major open question is whether it is possible to construct practical LAMs that are computationally efficient, safe, and explainable without the resource-intensive training of massive new sequence models.

\par In this work, we evaluate whether competent LAM intelligence can be achieved by the composition of compute-efficient smaller models, where action verification is driven by symbolic logic. Specifically, this paper offers the following contributions:

\begin{itemize}
  \item \textbf{We describe a modular architecture that achieves LAM capabilities by augmenting LLMs with extant perception models.} By integrating pre-trained vision models to provide the LLM with grounded perception, we demonstrate a method to build competent action models without the need for additional training. This approach not only avoids the costs of end-to-end training but also make the system's perceptual capabilities more extendable.
  
  \item \textbf{We introduce and evaluate two distinct planning frameworks: a purely neural approach and a neuro-symbolic approach.} In the \textit{neuro-symbolic} approach, we mitigate the risk of invalid PDDL generation by restricting the LLM's role. We rely on the LLM only for dynamic state generation at runtime, while static elements (such as the robot's actions and capabilities) are hard-coded deterministically. This hybrid PDDL is then solved by a traditional solver to ensure plan correctness and reduce hallucinations. In the \textit{neural} approach, we enforce safety by preventing the LLM from directly generating executable robot code. Instead, we utilize a hierarchical structure where a medium-level planner maps high-level subtasks to deterministic tools based on the robot's specific capabilities.

  \item \textbf{We implement a human-in-the-loop verification stage for both planning workflows.} We demonstrate an interface where the generated plan is presented to the user for review and revision prior to execution, ensuring transparency and interpretability.

  \item \textbf{We validate these hypotheses through a prototype engineering assistant.} We provide experimental evidence on a Universal Robots (UR) arm, demonstrating that this composite approach successfully grounds natural language commands into safe, interpretable physical actions.
\end{itemize}

\par The remainder of this paper is organized as follows: Section \ref{sec:arch} delineates the proposed modular architecture and the ``symbolic wrapping'' methodology designed to enforce control over Large Action Models. Section \ref{sec:implementation} details the practical implementation of this framework within a ROS2-based robotic arm system. Section \ref{sec:experiments} presents our experimental validation, offering a comparative analysis between neural-direct and neuro-symbolic planning strategies regarding success rates and safety. Section \ref{sec:related-work} places our contribution in the context of related work on neural and hybrid planning. Finally, Section \ref{sec:conclusion} concludes the study and discusses future research directions.

\section{Composing Large Action Model Architectures for Intelligent Robots}\label{sec:arch}

\subsection{LAM componentization}

\begin{figure}[htbp]
    \centering
    \includegraphics[width=\linewidth]{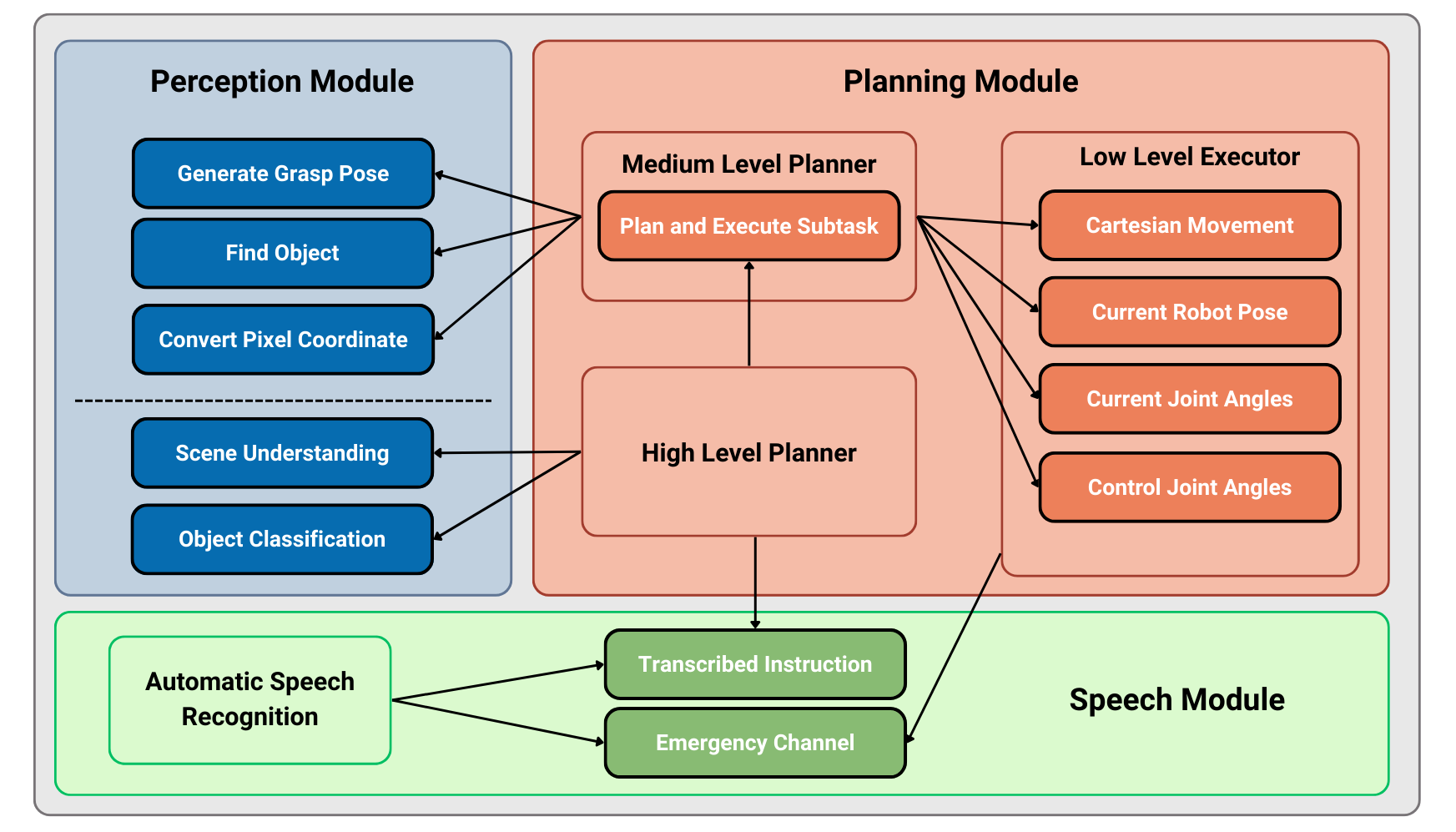}
    \caption{System Architecture Overview. The framework connects high-level cognitive reasoning with deterministic motion control through a stratified planning hierarchy.}
    \label{fig:architecture_overview}
    %\Description{System Architecture Overview}{System Architecture Overview. The framework connects high-level cognitive reasoning with deterministic motion control through a stratified planning hierarchy.}
\end{figure}

We propose a modular, neuro-symbolic architecture that functions as a comprehensive Large Action Model (LAM). Rather than relying on a single, opaque neural network, our system is composed of specialized functional modules that integrate perception, reasoning, and action. As illustrated in Figure~\ref{fig:architecture_overview}, the architecture is defined by a hierarchical planning pipeline driven by multi-modal inputs.

The system's awareness is driven by two multi-modal input modules. The \textit{Perception Module} employs a variety of open-vocabulary foundation models to perform object segmentation, classification, and grasp synthesis. This converts raw pixel data into a useful information without requiring task-specific training. Simultaneously, the \textit{Speech Module} utilizes a neural speech-to-text engine to capture user intent. This interface includes a symbolic override mechanism, where specific safety keywords (e.g., ``Stop'') bypass the reasoning layers to trigger immediate hardware halts.

\par To bridge the gap between abstract natural language and precise robot motor control, we implement a three-tier planning hierarchy. This hierarchical planning core ensures that high-level reasoning is grounded in valid physical capabilities.

\begin{itemize}
    \item \textbf{High-Level Reasoning (The Cognitive Agent):} At the top of the hierarchy, the system interprets user commands using two distinct paradigms: a \textit{neural} approach, where an LLM directly generates action sequences for flexibility, and a \textit{neuro-symbolic} approach, where the LLM translates requests into formal problem definitions (e.g., PDDL) for verification by a symbolic solver.
    
    \item \textbf{Medium-Level Orchestration (The Translator):} This layer acts as an intermediary, treating the robot's capabilities as callable ``tools.'' It decomposes the high-level subtasks into specific action sequences if needed. This layer manages the flow of execution and facilitates dynamic feedback between the planner and the perception modules.
    
    \item \textbf{Low-Level Execution (The Controller):} The foundation of the hierarchy is the motion control core. It handles path planning, trajectory generation, and collision avoidance. This layer exposes interfaces for Cartesian and joint-space control, ensuring that the abstract plans received from above are executed with kinematic precision and adherence to safety constraints.
\end{itemize}

This stratified design allows the robot to benefit from the flexibility of modern Generative AI while maintaining the reliability and safety essential for physical interaction.

\subsection{Symbolic Wrapping of LAMs}

\begin{figure}[htbp]
    \centering
    \includegraphics[width=\linewidth]{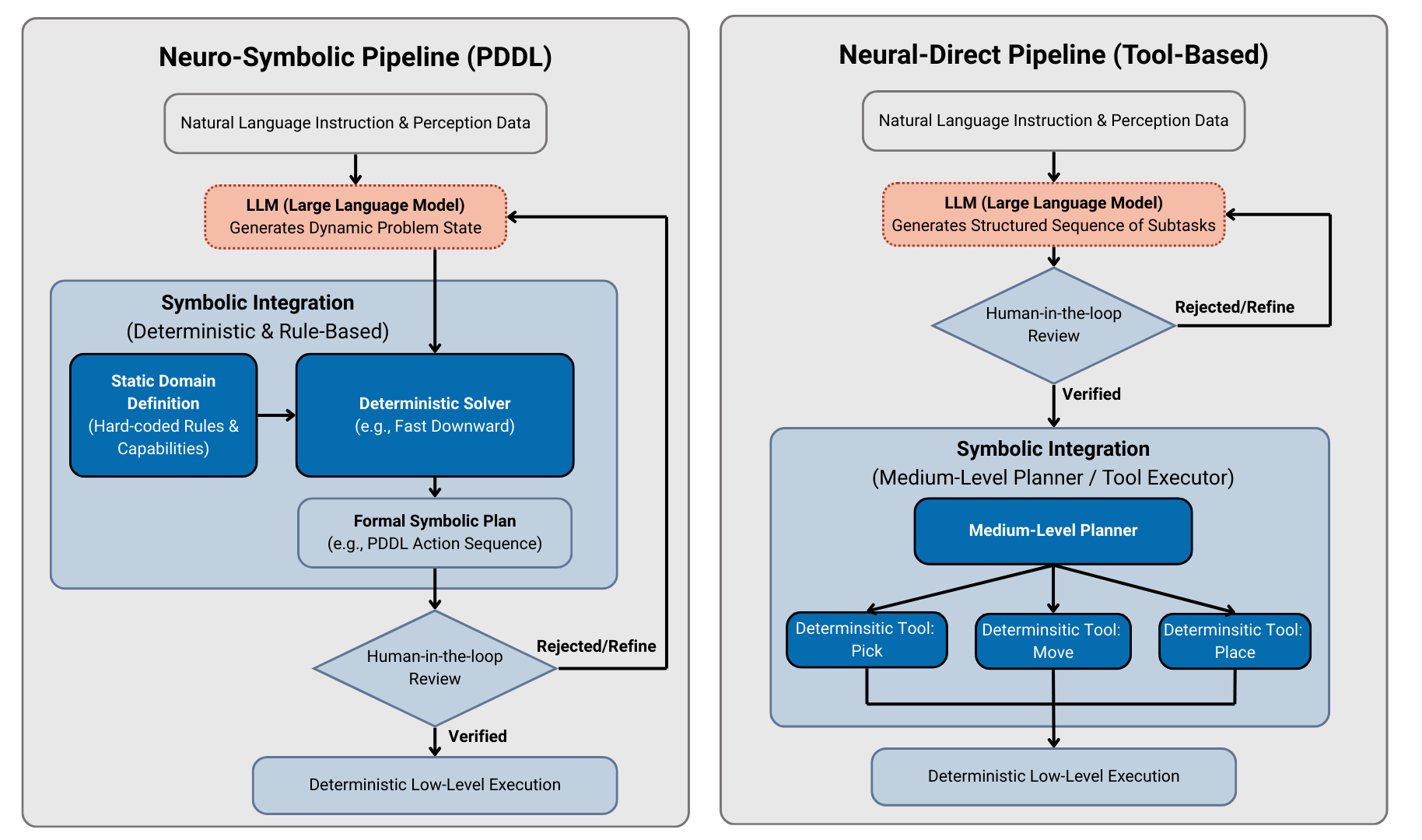}
    \caption{Overview of Symbolic Wrapping Architecture. The diagram illustrates how stochastic Large Language Models are insulated by deterministic symbolic layers. Both the neuro-symbolic and neural-direct pipelines generate an intermediate, verifiable symbolic artifact before any physical execution is attempted, enabling logic-based verification and human-in-the-loop intervention.}
    \label{fig:symbolic_wrapping}
    %\Description{Overview of Symbolic Wrapping Architecture}{Overview of the Symbolic Wrapping Architecture. The diagram illustrates how stochastic Large Language Models are insulated by deterministic symbolic layers. Both the neuro-symbolic and neural-direct pipelines generate an intermediate, verifiable symbolic artifact before any physical execution is attempted, enabling logic-based verification and human-in-the-loop intervention.}
\end{figure}

To address the lack of interpretability and control in Large Action Models, we implement a "symbolic wrapping" strategy. This approach restricts the Large Language Model (LLM) from generating executable robot code directly. Instead, the model is constrained to produce intermediate symbolic representations, structured plans or formal logic definitions, which act as a verification layer before any physical motion occurs. An overview of this architecture, illustrating the workflows for both pipelines, is depicted in Figure~\ref{fig:symbolic_wrapping}.

In our neuro-symbolic pipeline, we reduce the risk of invalid symbolic outputs by decoupling the domain definition from the problem instance. The robot's constraints and capabilities are hard-coded in a static symbolic framework, and the robot's current state is deterministically determined, while the LLM generates the dynamic problem state at runtime. By feeding this hybrid representation into a deterministic solver, we ensure that the output is not just syntactically correct, but logically solvable. If the LLM "hallucinates" an impossible goal, the solver fails to find a plan, effectively trapping the error in the software layer rather than causing a physical failure.

Similarly, in the neural-direct pipeline, we replace arbitrary code generation with structured tool usage. The LLM is restricted to selecting from a finite set of deterministic, pre-verified APIs. This limits the action space to safe operations and ensures that the model's intent is captured in a discrete list of sub-tasks.

This intermediate symbolic layer is critical for debugging and verification. Because the output is a structured interpretable artifact (a plan or task list) rather than opaque neural weights or raw code, it allows for a "human-in-the-loop" review process where operators can confirm or revise the plan before actual execution.

\section{Case Study: a Human-in-the-Loop Intelligent Robotic Arm}
\label{sec:implementation}

\begin{figure}[htbp]
\centering
\includegraphics[width=\linewidth]{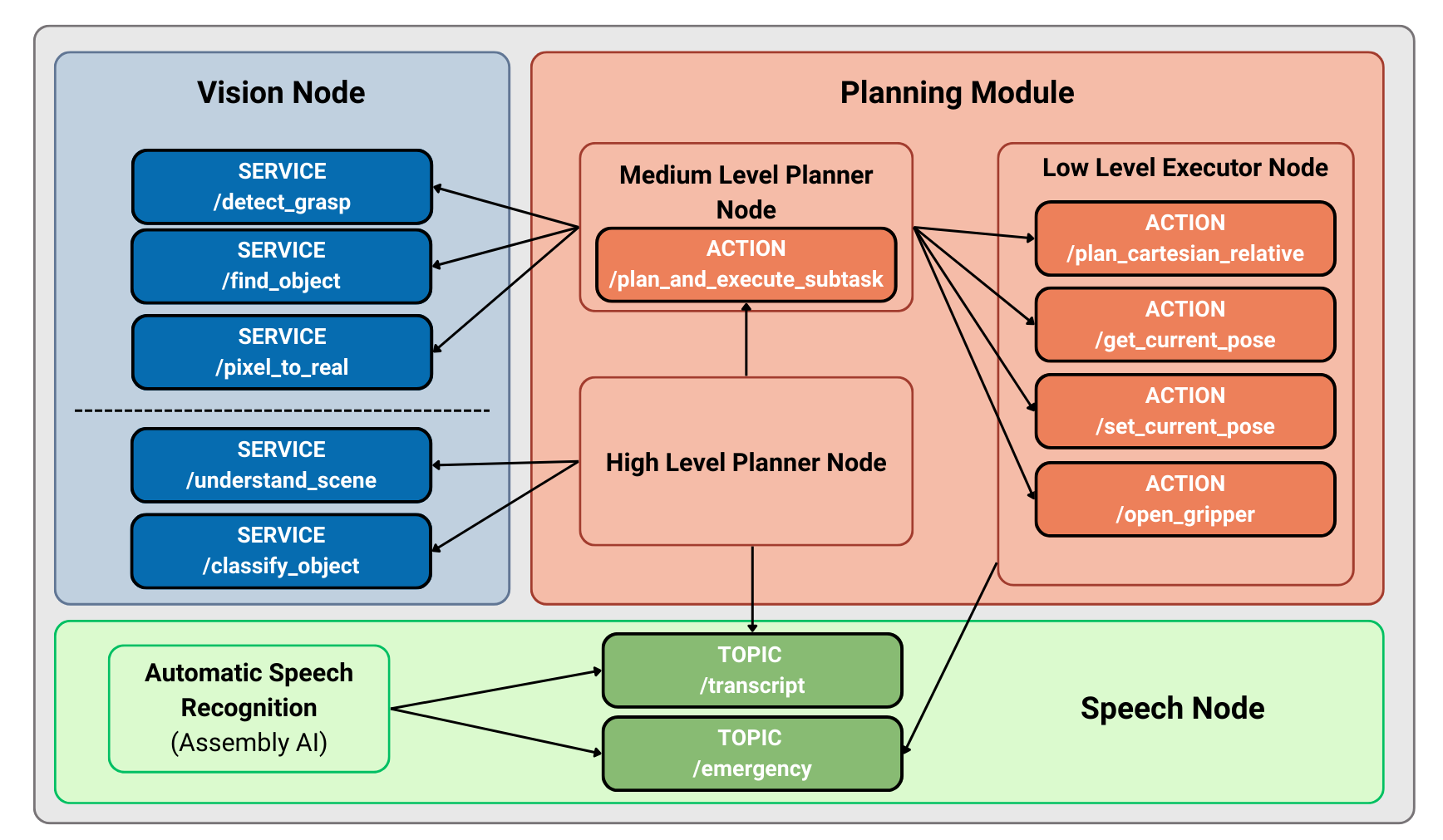}
\caption{ROS2 Graph for the Overall System. The system integrates perception services, event-driven speech topics, and action-based planning nodes.}
\label{fig:planning-result-ros2-graph}
%\Description{ROS2 Graph for the Overall System}{ROS2 Graph for the Overall System. The system integrates perception services, event-driven speech topics, and action-based planning nodes.}
\end{figure}

To validate the feasibility of constructing transparent and competent Large Action Models (LAMs) from extant building blocks, we developed a prototype fully integrated within the Robot Operating System 2 (ROS2) framework. As illustrated in Figure~\ref{fig:planning-result-ros2-graph}, the architecture is defined by a modular, hierarchical planning pipeline where each functional component operates as an independent ROS2 Node.

\textbf{Multi-modal Inputs.}
The system's awareness is driven by two dedicated ROS2 nodes that interface with the planning core through distinct communication patterns. The \textit{Perception Node} encapsulates open-vocabulary foundation models to perform object segmentation and classification. This node operates several ROS2 services, providing on-demand semantic data to the planning module only when queried. Conversely, the \textit{Speech Node} utilizes a neural speech-to-text engine to capture user intent and publishes transcribed text via ROS2 topics. This event-driven approach allows for asynchronous commands and includes a high-priority channel for safety keywords, which can trigger immediate system halts.

\textbf{Hierarchical Planning Core.}
To bridge the gap between abstract natural language and precise robot motor control, the planning module is distributed across three distinct ROS2 nodes. Intra-module communication is primarily handled via ROS2 Actions, allowing for long-running processes with feedback mechanisms.

\subsection{Perception Module}
Functioning as the sensory cortex of the system, the Perception Module bridges the gap between raw optical data and the \textit{Hierarchical Planning Core}. While the high-level planner requires semantic labels (e.g., ``red cube'') to reason about task logic, the low-level executor requires precise geometric affordances (e.g., 6-DoF poses) for actuation. Consequently, the perception architecture is designed to effectuate simultaneous geometric estimation and semantic reasoning through a service-oriented approach.

\par To ensure modularity and asynchronous operation, the vision pipeline is effected as a fully distributed \textbf{ROS2} architecture~\cite{macenski2022robot}. Utilizing \texttt{rclpy} for node orchestration and \texttt{cv\_bridge} for efficient image transport, the system decouples heavy inference tasks from the control loop. This design allows specific perception capabilities-segmentation, classification, grasping, or scene reasoning-to be invoked strictly on-demand by the Medium or High-Level Planners, minimizing computational overhead during idle states.

The perception stack comprises five specialized nodes, categorized by their contribution to the planning hierarchy:

\begin{enumerate}
    \item \textbf{Geometric Perception (Action-Centric):}
    \begin{itemize}
        \item \textit{SAM Detector} (\texttt{simple\_sam\_detector}): Leverages the \\Segment Anything Model (SAM)~\cite{kirillov2023segment} to generate instance masks and bounding boxes from raw RGB inputs.
        \item \textit{GraspNet Detector} (\texttt{graspnet\_detector}): Computes 6-DoF grasp configurations for detected instances using the GraspNet-1Billion baseline~\cite{fang2020graspnet}, publishing quality scores ($Q$) and approach vectors essential for the Low-Level Executor.
        \item \textit{Pixel-to-Real Service} (\texttt{pixel\_to\_real}): A utility node that continuously subscribes to depth streams, converting 2D pixel coordinates $(u, v)$ into 3D world coordinates $(x, y, z)$ via the camera's intrinsic matrix.
    \end{itemize}
    
    \item \textbf{Semantic Perception (Reasoning-Centric):}
    \begin{itemize}
        \item \textit{CLIP Classifier} (\texttt{clip\_classifier}): Performs zero-shot semantic classification on SAM-generated regions using the CLIP architecture~\cite{radford2021learning}. This provides the High-Level Planner with open-vocabulary object identification capabilities.
        \item \textit{Scene Understanding} (\texttt{scene\_understanding}): Derives a relational graph (e.g., adjacency, support, containment) from spatial data, allowing the planner to query environmental context (e.g., ``is object A to the left of object B or behind it?'').
    \end{itemize}
\end{enumerate}

\begin{figure}[htbp]
\centering
\includegraphics[width=\linewidth]{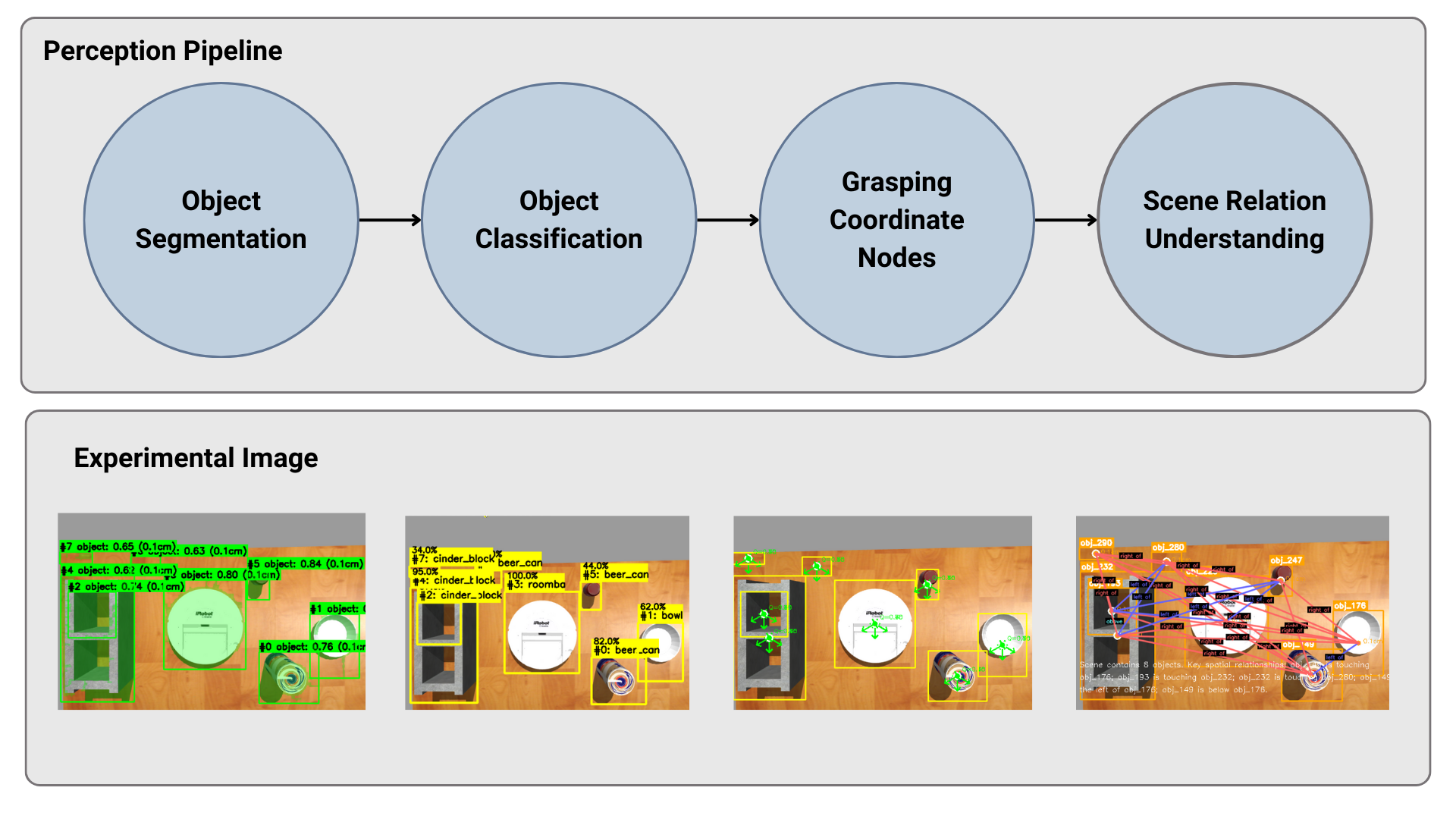}
\caption{The modular ROS2 perception architecture. Raw RGB-D data is processed into structured semantic and geometric messages, accessible to the planner via specific service calls.}
%\Description{The modular ROS2 perception architecture}{The modular ROS2 perception architecture. Raw RGB-D data is processed into structured semantic and geometric messages, accessible to the planner via specific service calls.}
\label{fig:perception_pipeline}
\end{figure}

\subsection{Speech Module}

To establish a reliable Automatic Speech Recognition (ASR) foundation for robot manipulation, a fully operational real-time speech-to-text system has been developed and deployed as a ROS2 node. The system continuously captures live microphone streams, transcribes them in real time, and publishes the recognized text for downstream processing on the UR-ARM. AssemblyAI functions as the primary transcription backend, utilizing advanced deep learning models to convert speech into text through a structured pipeline consisting of feature extraction, acoustic modeling, language modeling, and decoding \cite{assemblyai-docs}. This architecture enables robust handling of diverse speaking styles and environmental noise. Additional post-processing steps---such as punctuation insertion, capitalization, and disfluency removal---further improve the clarity and readability of the final output.

\begin{figure}[htbp]
    \centering
    \includegraphics[width=\linewidth]{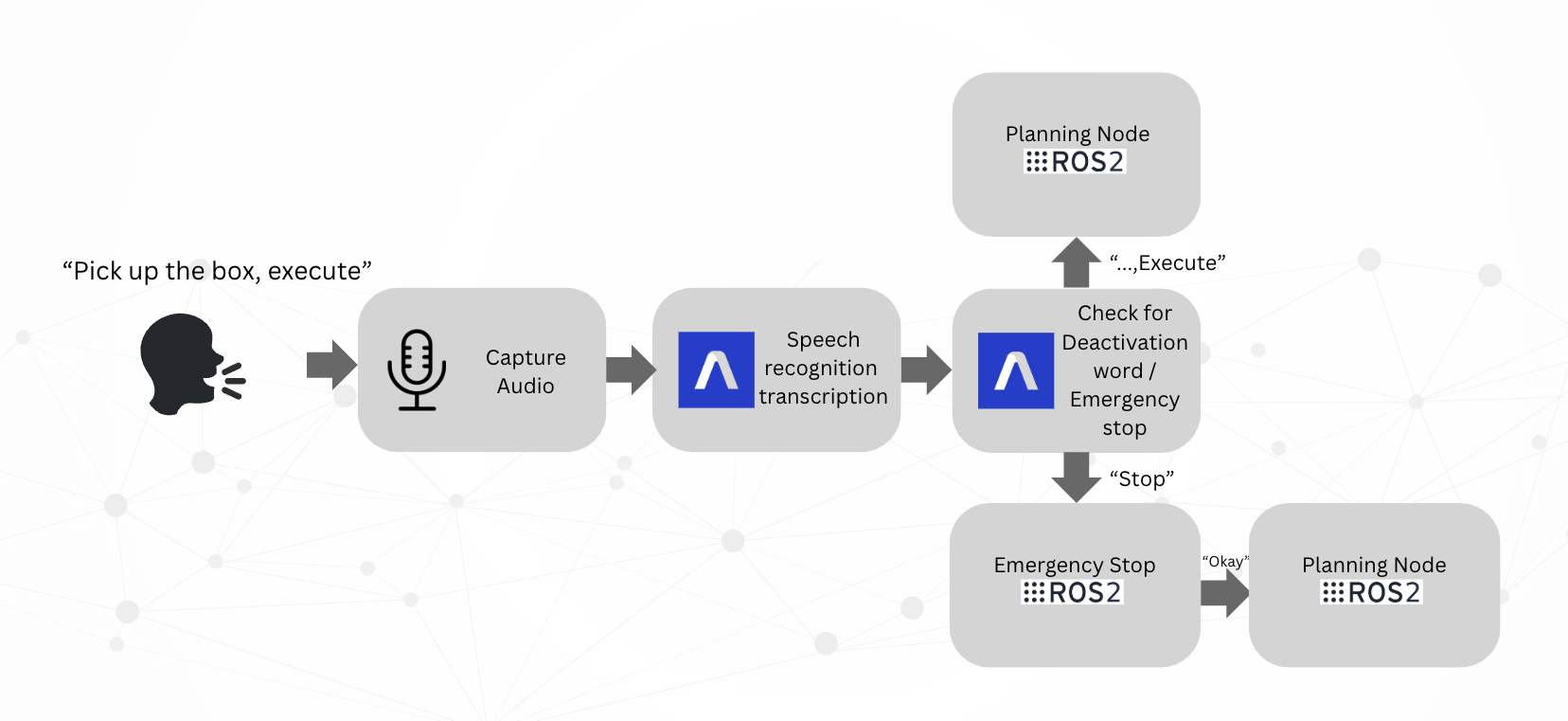}
    \caption{Overview of the real-time ASR pipeline for robot manipulation.}
    \label{fig:asrpipe.png}
    %\Description{Overview of the real-time ASR pipeline for robot manipulation.}
\end{figure}

As shown in Figure~\ref{fig:asrpipe.png}, the ASR node operates in a continuous cycle: capturing live audio, submitting it as a transcription request, polling for recognition results, and monitoring for additional user input. The system also incorporates a deactivation-word mechanism, sending commands to the planning node only when a spoken instruction ends with the keyword ``execute.'' An emergency-stop function is included as well; the robot halts immediately when the word ``STOP'' is detected and resumes only when the user says ``OKAY.'' These mechanisms replace the earlier three-second silence-based activation method, resulting in more responsive and deliberate user interactions.

This workflow provides a modular and scalable communication interface that supports safe, responsive, and context-aware interaction between humans and robots. The ASR module has been successfully integrated with the physical UR-ARM platform, establishing a foundation for future enhancements such as verbal confirmation feedback and task-completion notifications to further improve human--robot collaboration.

\subsection{Planning Module}

The implementation of the planning hierarchy relies on a synthesis of Large Language Model frameworks and standard robotics motion planning libraries.

\textbf{High-Level Planner.}
The reasoning agent is implemented using the LangChain framework~\cite{langchain-docs}, which manages the LLM context and tool usage. We explored two distinct implementations for this agent:
\begin{enumerate}
    \item \textbf{Neural Approach (LLM-Direct):} In this configuration, the LangChain agent is prompted to function as an end-to-end planner. It directly generates a structured sequence of sub-tasks based on the user's natural language instruction and the current perception context. 
    \item \textbf{Neuro-Symbolic Approach (PDDL-based):} Here, the LangChain agent functions strictly as a semantic translator. Instead of generating the plan directly, it interprets the natural language instruction to generate a symbolic problem file in the Planning Domain Definition Language (PDDL)~\cite{pddl-1.2}. This problem file, combined with a static domain definition, is then solved by Fast Downward~\cite{the-fast-downward-planning-system}, a deterministic symbolic planner. This ensures that the resulting plan is mathematically verifiable and logically sound before it is passed down the hierarchy.
\end{enumerate}

\textbf{Medium-Level Planner.}
The orchestration layer is also implemented as a LangChain agent. However, unlike the high-level planner, its role is deterministic execution management. It utilizes custom tools that act as wrappers around ROS2 service and action calls. This agent receives the sub-tasks (from either the direct LLM output or the Fast Downward solution) and translates them into specific API calls (e.g., triggering the \texttt{/plan\_cartesian\_relative} action), handling the immediate feedback loop with the robot's state.

\textbf{Low-Level Planner.}
The execution layer is handled by MoveIt2~\cite{moveit2}, which serves as the backend for path planning and trajectory generation. The \texttt{low\_level\_planner\_executor} node leverages MoveIt2's planning pipelines to solve for inverse kinematics and collision avoidance. It enforces Cartesian motion constraints and joint-space limits to ensure safe operation. By isolating the kinematic complexity within MoveIt2, the higher-level LangChain agents can focus entirely on task logic and semantic reasoning.

\section{Experiments and Results}
\label{sec:experiments}

To validate the hypothesis that competent Large Action Models (LAMs) can be constructed using a composition of third-party smaller components and symbolic wrappers, we conducted a series of experiments evaluating the system's modularity, planning capabilities, and safety response. The experimental framework was established across two environments, simulation and physical hardware, to ensure both scalability and physical applicability.

\subsection{System Validation and Environment}

\begin{figure}[htbp]
    \centering
    \includegraphics[width=0.9\linewidth]{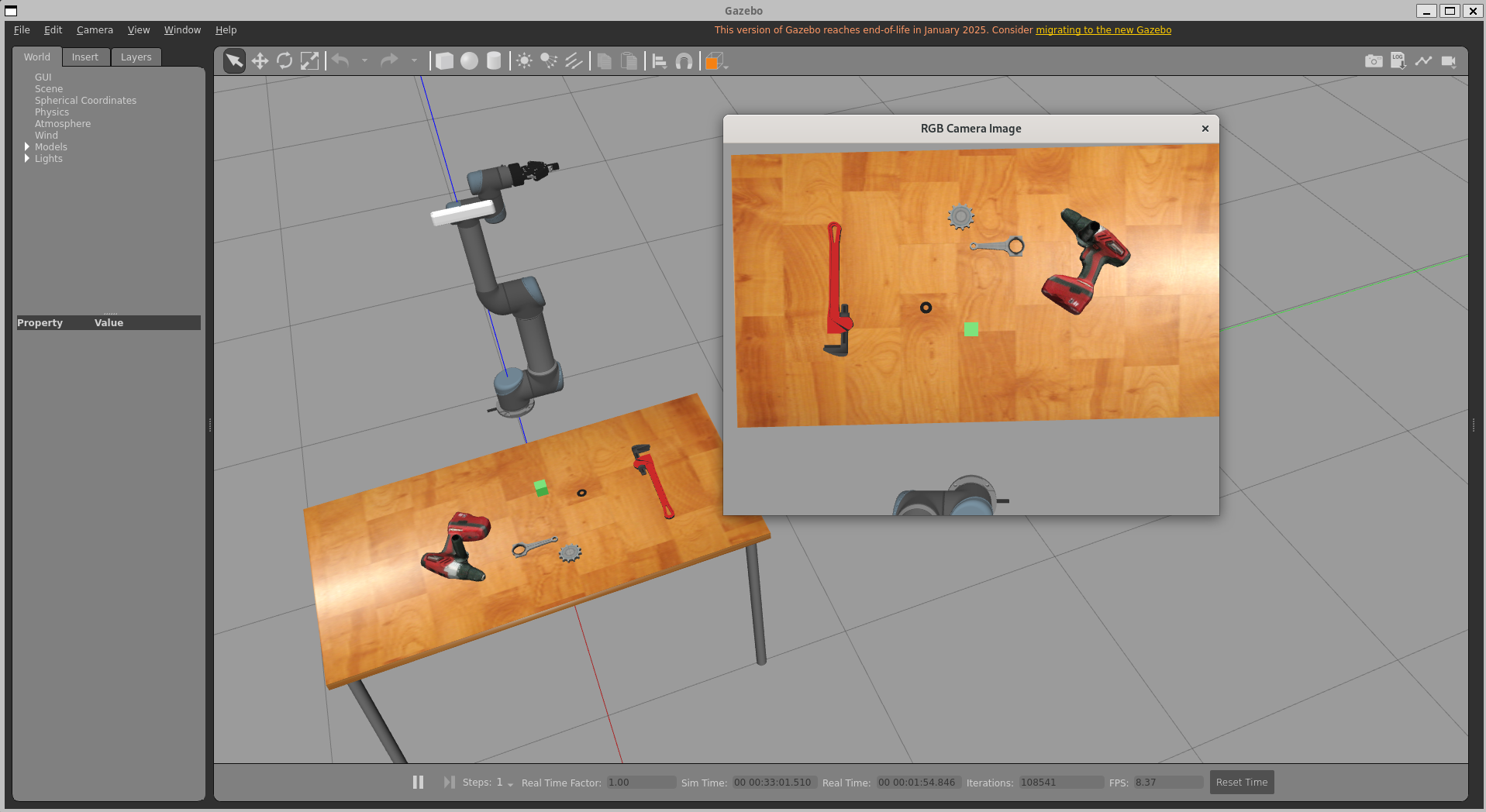}
    \caption{Simulation environment in Gazebo utilizing standard ROS2 interfaces for component validation.}
    %\Description{Simulation environment in Gazebo}{Simulation environment in Gazebo utilizing standard ROS2 interfaces for component validation.}
    \label{fig:simulation-gazebo-setup}
\end{figure}

The architecture was first deployed in a simulation using \textbf{Gazebo Classic}~\cite{gazebo}, modeling a UR5 robotic arm with a Robotiq 2F-85 gripper and depth perception (Figure~\ref{fig:simulation-gazebo-setup}). This environment validated the ROS2 integration graph and the perception-action loop prior to physical deployment.

\begin{figure}[htbp]
    \centering
    \includegraphics[width=0.9\linewidth]{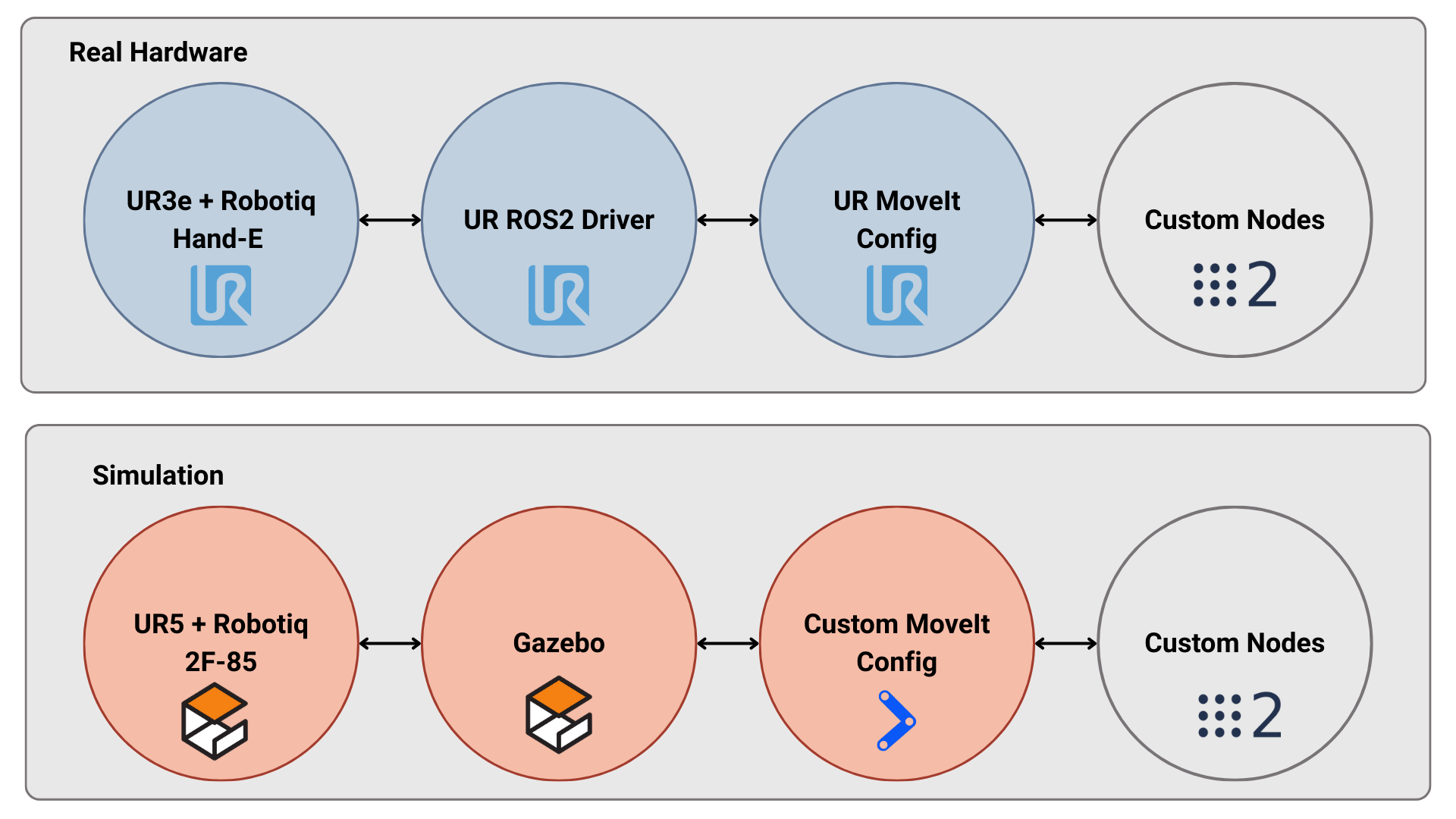}
    \caption{Comparison between Real Hardware Setup (UR3e with Robotiq HandE gripper) and Simulation (Gazebo).}
    %\Description{Comparison between Real Hardware and Simulation}{Comparison between Real Hardware Setup (UR3e with Robotiq HandE gripper) and Simulation (Gazebo).}
    \label{fig:hardware-result-sim-vs-real}
\end{figure}

Subsequently, the system was transferred to real hardware using a \textbf{UR3e robot} with a Robotiq HandE gripper. By leveraging the official UR ROS2 Driver~\cite{ur_ros2_driver} and unified ROS2 interfaces, the transition from simulation to reality required no architectural changes, confirming the portability of the component-based design (Figure~\ref{fig:hardware-result-sim-vs-real}).

Qualitative performance was established through a series of demonstrations illustrating the system's ability to interpret natural language, revise plans dynamically, and execute safety protocols:
\begin{itemize}
    \item \textbf{Full System Integration:} A demonstration of the complete pipeline (ASR, Planning, Perception) in Gazebo. The robot executes motion commands based on specific objects detected in the scene by the vision module. \\
    \href{https://drive.google.com/file/d/1draHgIV_FaXqrEH4p6pD_vMLNN0HLVL3/view?usp=sharing}{[Video Demonstration]}
    
    \item \textbf{Sim-to-Real Consistency:} A side-by-side comparison of the ASR and planning modules driving both the simulated robot and the physical UR3e simultaneously, highlighting the architectural uniformity. \\
    \href{https://drive.google.com/file/d/1yYsGxQITKcc62LWY9pjuGWzi5-gLacgh/view?usp=sharing}{[Video Demonstration]}
    
    \item \textbf{Dynamic Plan Revision:} A demonstration of the system's interpretability and memory. The user interrupts an initially generated plan with a natural language correction (``Swap the action order''), and the system correctly regenerates the task sequence before execution. \\
    \href{https://drive.google.com/file/d/1QjeXoCuzd0tnjiekhAuwajCcLh3JXEpf/view?usp=sharing}{[Video Demonstration]}
\end{itemize}

\subsection{Perception Module}

The vision pipeline was rigorously evaluated in a high-fidelity Gazebo simulation environment to establish baseline performance metrics prior to physical deployment.

The evaluation dataset consisted of 5 distinct world configurations, each populated with 5 randomized objects, totaling 25 unique object instances. The testing protocol involved an end-to-end execution where the \texttt{/vision/run\_pipeline} service was triggered for each scene. We logged outputs from all sub-modules and compared them against ground-truth simulation states.

Performance was quantified using standard computer vision metrics adapted for robotic tasks:
\begin{itemize}
    \item \textbf{Segmentation:} Average Precision (AP) at Intersection-over-Union (IoU) $\ge 0.5$.
    \item \textbf{Classification:} Top-1 Accuracy, measuring the fraction of proposals where the highest-scoring CLIP label matched the ground truth.
    \item \textbf{Grasp Quality:} Mean Quality Score ($Q$), a consolidated metric of grasp feasibility (antipodal geometry and collision clearance).
    \item \textbf{Spatial Accuracy:} The success rate of the scene understanding module in correctly identifying relative spatial predicates (e.g., \texttt{left-of}, \texttt{on-top-of}) and metric positions within a 2cm tolerance.
    \item \textbf{Metric Projection:} Root Mean Square Error (RMSE) of the pixel-to-real coordinate conversion.
\end{itemize}

\textbf{Results.}
Table~\ref{tab:vision-eval} summarizes the aggregate performance across all trials. The system demonstrated high geometric fidelity, with the Pixel-to-Real service achieving an RMSE of $0.054$m, well within the capture tolerance of standard parallel-jaw grippers. The Scene Understanding module achieved a $0.98$ success rate, validating the High-Level Planner's ability to construct reliable mental models of the environment.

\begin{table}[htbp]
\centering
\caption{Aggregate perception module benchmark results ($N=25$ Objects).}
\label{tab:vision-eval}
\begin{tabular}{lcc}
\hline
\textbf{Module} & \textbf{Metric} & \textbf{Result} \\
\hline
SAM (Segmentation) & AP (IoU $\ge$ 0.5) & $0.741$ \\
CLIP (Classification) & Top-1 Accuracy & $0.850$ \\
GraspNet (Pose Estimation) & Mean Quality Score ($Q$) & $0.746$ \\
Scene Understanding & Spatial Accuracy & $0.980$ \\
Pixel-to-Real Service & RMSE (meters) & $0.054$ \\
\hline
\end{tabular}
\end{table}

While GraspNet achieved a respectable mean quality score of $0.746$, failures were primarily observed in scenarios with heavy occlusion or extreme object overlap. However, the high classification accuracy ($85\%$) suggests that the semantic layer is sufficiently robust to support natural language grounding in the subsequent planning phases.

\subsection{Speech Module}

With the ASR system fully integrated as a ROS2 node for the UR-ARM, enabling real-time speech-to-text processing, deactivation-word detection, and an emergency stop mechanism. The node continuously captures live audio, transcribes commands, and forwards them to the planning node only when the utterance ends with the deactivation word ``execute.'' The emergency stop is triggered by the word ``STOP'' and remains active until the user says ``OKAY.'' This design ensures responsive, safe, and reliable voice-based interaction between the operator and the robot.

The system was evaluated using 30 task-related voice commands of varying complexity. Three participants executed all commands. Performance was measured by successful recognition and correct transmission to the planning node.

\begin{table}[t]
\centering
\small
\caption{ASR Command Recognition Success for Three Participants}
\resizebox{\textwidth}{!}{%
\begin{tabular}{lccc}
\hline
\textbf{Sentences} & \textbf{Participant 1} & \textbf{Participant 2} & \textbf{Participant 3} \\
\hline
% --- table content unchanged ---
Pick up the red cube and place it on the table, execute. & 12/12 & 12/12 & 12/12 \\
Move the arm to the home position, execute. & 8/8 & 8/8 & 8/8 \\
Grab the green cylinder and put it in the bin, execute. & 11/11 & 11/11 & 11/11 \\
Rotate the wrist clockwise 90 degrees, execute. & 7/7 & 7/7 & 7/7 \\
Pick up the yellow object and stack it on the red cube, execute. & 13/13 & 13/13 & 13/13 \\
Extend the arm forward slowly, execute. & 6/6 & 6/6 & 6/6 \\
Move to the left side of the table, execute. & 9/9 & 9/9 & 9/9 \\
Stop. & 1/1 & 1/1 & 1/1 \\
Okay. & 1/1 & 1/1 & 1/1 \\
Lift the blue box and place it on the top shelf, execute. & 12/12 & 12/12 & 12/12 \\
Move the arm to the scanning position above the center point, execute. & 12/12 & 12/12 & 12/12 \\
Pick up the small screw and hand it to me on the right side, execute. & 14/15 & 15/15 & 14/15 \\
Lower the object two centimeters and hold position, execute. & 9/9 & 9/9 & 8/9 \\
Slide the object forward and align it with the edge, execute. & 11/11 & 11/11 & 11/11 \\
Place the green block next to the yellow block on the left, execute. & 13/13 & 13/13 & 13/13 \\
Rotate the gripper counterclockwise 45 degrees, execute. & 7/7 & 7/7 & 7/7 \\
Pick up the blue cylinder and rotate it vertically, execute. & 10/10 & 9/10 & 10/10 \\
Lift the object slightly and center it over the target marker, execute. & 12/12 & 12/12 & 12/12 \\
Move the arm upward until it reaches the safe height, execute. & 11/11 & 11/11 & 11/11 \\
Pick up the tool and position it in front of the camera, execute. & 13/13 & 13/13 & 12/13 \\
Push the red cube gently toward the right side, execute. & 10/10 & 9/10 & 9/10 \\
Reach toward the far corner and check for any obstacles, execute. & 11/11 & 11/11 & 10/11 \\
Pick up the cup and place it inside the container, execute. & 11/11 & 11/11 & 11/11 \\
Pick up the screwdriver and place it in the tool holder slot, execute. & 13/13 & 13/13 & 13/13 \\
Move the arm in a straight line back to the origin point, execute. & 13/13 & 13/13 & 13/13 \\
Pick up the object and align it with the marked orientation, execute. & 12/12 & 11/12 & 12/12 \\
Grasp the black box and slide it to the center, execute. & 11/11 & 10/11 & 11/11 \\
Rotate joint three to the forty-degree position, execute. & 9/9 & 9/9 & 9/9 \\
Lift the object, move it behind the container, and lower it gently, execute. & 13/13 & 13/13 & 13/13 \\
\hline
\textbf{Results} & 0.998 & 0.9875 & 0.9518 \\
\textbf{Average} & \multicolumn{3}{c}{0.9791} \\
\hline
\end{tabular}%
}
\end{table}

\noindent \textbf{Overall performance.}  
Across all participants, the ASR achieved a mean accuracy of $0.9791$.  
The standard deviation was $s = 0.0242$, and using a 95\% confidence interval (df = 2), the resulting interval was:

\[
0.9791 \pm 0.0602 \quad (95\%~\text{CI: } [0.919,\,1.000])
\]

\noindent (Upper bound capped at 1.000.)

The ASR system consistently recognized simple and short commands with perfect accuracy. Occasional mis-recognition occurred only in longer, multi-step instructions. Both the deactivation word (``execute'') and emergency-stop commands (``STOP,'' ``OKAY'') were detected reliably in all trials.

Overall, the results demonstrate that the ASR module provides accurate command interpretation, robust communication with the planning node, and reliable safety handling for real-time human--robot manipulation tasks.

\subsection{Planning Module}

The core hypothesis of this work suggests that competent Large Action Models should not rely on end-to-end generation of low-level control code, but rather on the composition of high-level reasoning with symbolic constraints. We evaluated two implementations of this philosophy:
\begin{enumerate}
    \item \textbf{LLM-Direct (Tool-Use):} The model maps instructions to a structured sequence of sub-tasks or actions which are then executed using a restricted set of pre-defined API "tools" (e.g., \texttt{move\_to}, \texttt{grip}), effectively wrapping neural outputs in a functional symbolic layer.
    \item \textbf{Neuro-Symbolic (PDDL):} The model translates instructions into a formal PDDL problem definition, which is solved deterministically by the Fast Downward planner.
\end{enumerate}

\textbf{Control and Interpretability:}
Crucially, both methods differ from "black-box" end-to-end models by enforcing a verification layer. Whether generating a sequence of tool calls or a PDDL problem instance, the system produces an intermediate, human-readable plan. This allows for the \textit{Plan Revision} behavior demonstrated in our experiments, where high-level plans can be inspected and modified by the user prior to any physical execution. This architecture ensures that the stochasticity of the LLM is confined to the reasoning phase, while the execution phase remains transparent and bounded by the robot's defined capabilities.

\textbf{Comparative Performance:}
We evaluated both approaches on a set of 13 tasks ($N=65$ total trials). The results (Table~\ref{tab:planning-eval}) indicate that while both methods successfully leverage symbolic grounding to control output, they trade off between flexibility and strictness.

The \textbf{LLM-Direct} approach achieved a 100\% success rate, utilizing its "tool-calling" capability to flexibly handle abstract instructions that are difficult to model in strict PDDL. The \textbf{Neuro-Symbolic} approach (91\% success) was slightly faster ($6.83$s vs $7.20$s) and provided rigorous mathematical guarantees on plan validity; however, it was more brittle when the LLM failed to generate syntactically perfect PDDL.

\begin{table}[h!]
    \centering
    \caption{Performance Comparison of Neural (LLM-Direct) and Neuro-Symbolic (PDDL-Based) Planning}
    \label{tab:planning-eval}
    \begin{tabular}{lcc}
    \hline
    \textbf{Metric} & \textbf{LLM-Direct} & \textbf{Neuro-Symbolic (PDDL)} \\
    \hline
    Avg. Execution Time per Step (s) & $7.20 \pm 0.25$ & $\mathbf{6.83 \pm 0.27}$ \\
    Success Rate (\%) & $\mathbf{100.0}$ & $91.0$ \\
    LLM Requests per Step & $2.0$ & $2.0$ \\
    Computational Cost (Tokens) & $\approx 3{,}000$ & $\approx 3{,}000$ \\
    \hline
    \end{tabular}
    \begin{flushleft}
    \footnotesize{Note: Time difference is statistically significant ($p = 0.049$).}
    \end{flushleft}
\end{table}

\textbf{Safety and Responsiveness:} A critical requirement for LAMs operating in human-centric spaces is the ability to override AI autonomy with deterministic safety layers. We measured the system's latency in processing an ``Emergency Stop'' voice command during active motion. The system achieved an average stop latency of $1.41 \pm 0.14$ seconds. This confirms that the hierarchical architecture successfully decouples high-level AI planning from low-level safety protocols, ensuring that human intervention remains preemptive regardless of the complexity of the current AI task.

\section{Related Work}
\label{sec:related-work}

The realization of intelligent robotic agents requires bridging the gap between high-level semantic understanding and low-level physical actuation. The literature describes a progression from purely symbolic planners to data-driven neural networks, and more recently, towards integrated Large Action Models (LAMs) and hybrid architectures.

\par\subsection{Classic Planning and Symbolic Control.}
Historically, robot planning has relied on the segregation of high-level task planning and low-level motion planning. Traditional methods utilize hand-coded symbols and strict hierarchical decompositions to map environment states to actions~\cite{learning-neuro-symbolic, code-as-symbolic-planner}. A cornerstone of this approach is the Planning Domain Definition Language (PDDL), which allows solvers to generate reliable, long-horizon plans based on pre-defined logical operators~\cite{optimizatoin-and-motion-planning}.
While these symbolic approaches offer high interpretability and guarantee that constraints are met, they face significant scalability barriers. They rely on fixed domain-specific languages, making them brittle in unstructured environments and difficult to scale to new tasks without expert intervention~\cite{ns-vqa}. Furthermore, they often struggle to effectively integrate discrete task decisions with the continuous geometric and kinematic constraints required for dynamic manipulation~\cite{tamp, cliport}. Despite these limitations, the explicit reasoning chain provided by symbolic planners remains a critical asset for verification and safety in robotic systems.

\par\subsection{Neural Planning and Vision-Language-Action Models.}
To address the rigidity of symbolic systems, recent research has pivoted toward neural approaches, specifically leveraging Large Language Models (LLMs) and Vision-Language-Action (VLA) models. LLM-based planners exploit the broad commonsense reasoning of pre-trained models to interpret natural language instructions. Approaches such as \textit{Code-as-Policies} query LLMs to generate executable Python code that creates robot actions from pre-defined primitives~\cite{code-as-policies}.
Recent advancements have focused on end-to-end VLA models that unify perception and control. For instance, GraspVLA introduces a pre-training paradigm relying entirely on large-scale synthetic action data. It utilizes a Progressive Action Generation mechanism, treating perception tasks as intermediate Chain-of-Thought steps to generate continuous robot actions, thereby bridging the sim-to-real gap without costly real-world data collection~\cite{graspvla}.
However, these monolithic neural approaches often operate as "black boxes," suffering from hallucination, a lack of explainability, and difficulties in handling complex numerical constraints~\cite{vla-in-robot}. While they excel at open-world visual grounding, they typically lack the formal verification mechanisms necessary for safe deployment in collaborative environments~\cite{plangenllm}.

\par\subsection{Large Action Models and Hybrid Architectures.}
The convergence of these paradigms has led to the emergence of Large Action Models (LAMs), designed to actively interact with and manipulate physical and digital environments rather than just generating text~\cite{lam}. For example, the xLAM series proposes a family of models specifically optimized for autonomous agent tasks, such as function calling and multi-turn interaction, achieving state-of-the-art performance by unifying and synthesizing diverse agent datasets~\cite{xlam}.
To ensure control and reliability, newer approaches are moving away from purely monolithic designs toward structured, neuro-symbolic, or hierarchical architectures. DAHLIA, for instance, introduces a data-agnostic framework employing a "dual-tunnel" architecture. It pairs an LLM-powered planner with a reporter module that provides closed-loop feedback, utilizing temporal abstraction and Chain-of-Thought reasoning to enable adaptive re-planning and failure recovery in long-horizon tasks~\cite{dahlia}. Similarly, recent studies suggest that modern recurrent architectures, such as xLSTM, can serve as efficient backbones for Large Recurrent Action Models (LRAMs), offering linear-time inference complexity suitable for real-time robotics tasks~\cite{lram}.
These structured approaches disentangle perception from reasoning, using neural networks for visual grounding and symbolic or structured engines for planning~\cite{nsai, enhancing-interpret}. By verifying neural outputs against constraints, such as those defined in PDDL, these hybrid frameworks mitigate the "black box" problem, offering a pathway to robust, interpretable, and verifiable Large Action Models~\cite{audere}.

\section{Conclusions}
\label{sec:conclusion}

\par
This work demonstrates that the capabilities of Large Action Models (LAMs) can be effectively realized through a modular, neuro-symbolic architecture rather than relying exclusively on massive end-to-end training. We establish that by composing off-the-shelf perception models with a logic-driven core, it is possible to build intelligent robotic systems that remain interpretable and controllable. Specifically, we show that wrapping stochastic Large Language Models with deterministic symbolic constraints, whether through structured tool use or formal PDDL generation, bridges the gap between open-ended reasoning and the rigorous safety requirements of physical actuation.

\par
Our experiments verify that this composite approach successfully grounds natural language commands into safe physical actions while enabling critical human-in-the-loop intervention. We found that while a neural-direct approach offers greater flexibility for abstract instructions (achieving a 100\% success rate in our trials), the neuro-symbolic PDDL approach provides superior mathematical guarantees on plan validity, albeit with increased sensitivity to syntax generation errors. Crucially, the decoupling of high-level reasoning from low-level control allowed for a safety override latency of less than 1.5 seconds, proving that agentic intelligence can be deployed without compromising hardware safety protocols. The lesson learned is that the generation of an intermediate, human-readable symbolic plan is essential for preventing the "action hallucinations" common in black-box models.

\par
Despite these advancements, the reliability of generating formal symbolic structures remains a bottleneck. Future research must address the brittleness of neuro-symbolic translation, particularly by developing methods for the iterative self-correction of generated PDDL code to match the flexibility of direct neural approaches. Furthermore, for these systems to scale beyond structured settings, the field must move towards dynamic domain generation, where the robot can autonomously expand its symbolic understanding of the world based on novel perceptual inputs, rather than relying on static, hand-coded domain definitions.

\bibliography{sample}

\section*{Acknowledgements}

The authors would like to thank the Robotics \& AI committee at the International School of Engineering, Chulalongkorn University, for financially supporting this project.

\end{document}